%% file: multitask_aggregated.tex
\documentclass{article}

\usepackage[final]{neurips_2019}

\usepackage[utf8]{inputenc} 
\usepackage[T1]{fontenc}    
\usepackage{hyperref}       
\usepackage{url}            
\usepackage{booktabs}       
\usepackage{amsfonts}       
\usepackage{nicefrac}       
\usepackage{microtype}      
\usepackage{amssymb}
\usepackage{amsmath}
\usepackage{bm}
\usepackage{graphicx}
\usepackage{subcaption}
\usepackage{caption}
\usepackage{float}

\DeclareMathOperator{\erf}{erf}

\DeclareMathOperator{\cov}{cov} 

\newcommand{\boldx}{\mathbf{x}} 
\newcommand{\boldz}{\mathbf{z}} 
\newcommand{\boldZ}{\mathbf{Z}} 
\newcommand{\boldy}{\mathbf{y}} 
\newcommand{\boldf}{\mathbf{f}} 
\newcommand{\boldu}{\mathbf{u}} 
\newcommand{\boldK}{\mathbf{K}} 
 
\newcommand{\boldB}{\mathbf{B}} 

\title{Multi-task Learning for Aggregated Data using Gaussian Processes}

\author{%
	Fariba Yousefi\\
	 \And
	 Michael Thomas Smith \\
	 \And
	 Mauricio A. \'Alvarez \\
	 \AND
	 \\
 	Department of Computer Science, University of Sheffield\\
 	\texttt{\{f.yousefi, m.t.smith, mauricio.alvarez\}@sheffield.ac.uk} \\
}

\begin{document}
\maketitle

\begin{abstract}
Aggregated data is commonplace in areas such as epidemiology and
demography. For example, census data for a population is usually
given as averages defined over time periods or spatial resolutions
(cities, regions or countries). In this paper, we present a novel
multi-task learning model based on Gaussian processes for joint learning of variables that
have been aggregated at different input scales. Our model represents
each task as the linear combination of the realizations of latent processes that are integrated
at a different scale per task. We are then able to compute the cross-covariance between the different tasks
either analytically or numerically. We also allow each task to have a potentially different likelihood model and
provide a variational lower bound that can be optimised in a
stochastic fashion making our model suitable for larger datasets. We
show examples of the model in a synthetic example, a fertility dataset and an air pollution
prediction application.
\end{abstract}

\section{Introduction}
Many datasets in fields like ecology, epidemiology, remote sensing,
sensor networks and demography appear naturally aggregated, that
is, variables in these datasets are measured or collected in intervals, areas or supports of different shapes and sizes. For example, census data are
usually sampled or collected as aggregated at different administrative
divisions, e.g. borough, town, postcode or city levels. In sensor networks, correlated variables are measured
at different resolutions or scales. In the near future, air pollution
monitoring across cities and regions could be done using a combination
of a few high-quality low time-resolution sensors and several low-quality (low-cost) high time-resolution sensors. Joint modelling of the variables registered in the census data or
the variables measured using different sensor configurations at different scales can improve predictions at the point or support levels.  

In this paper, we are interested in providing a general framework for
multi-task learning on these types of datasets. Our motivation is to
use multi-task learning to jointly learn models for different tasks
where each task is defined at (potentially) a different support of any
shape and size and has a (potentially) different nature, i.e. it is a
continuous, binary, categorical or count variable. We appeal to the
flexibility of Gaussian processes (GPs) for developing a prior over such
type of datasets and we also provide a scalable approach for
variational Bayesian inference.

Gaussian processes have been used before for aggregated data
\citep{Smith:binned:2018, Law:variational:aggregate:neurips:2018,
  tanaka2019refining} and also in the context of the related field of
\emph{multiple instance learning} \citep{Kim:MIL:GP:ICML:2010,
  Kandemir:MIL:GP:BMVC:2016,Haubmann:MIL:GP:CVPR:2017}. In multiple
instance learning, each instance in the dataset consists of a set (or
\emph{bag}) of inputs with only one output (or label) for that whole
set. The aim is to provide predictions at the level of individual
inputs. \citet{Smith:binned:2018} provide a new kernel function to
handle single regression tasks defined at different supports. They use
cross-validation for hyperparameter
selection. \citet{Law:variational:aggregate:neurips:2018} use the
weighted sum of a latent function evaluated at different inputs as the
prior for the rate of a Poisson likelihood.  The latent function
follows a GP prior. The authors use stochastic variational inference
(SVI) for approximating the posterior
distributions. \citet{tanaka2019refining} mainly use GPs for creating
data from different auxiliary sources. Furthermore, they only consider
Gaussian regression and they do not include inducing variables. While
\citet{Smith:binned:2018} and
\citet{Law:variational:aggregate:neurips:2018} perform the aggregation
at the latent prior stage, \citet{Kim:MIL:GP:ICML:2010,
  Kandemir:MIL:GP:BMVC:2016} and \citet{Haubmann:MIL:GP:CVPR:2017}
perform the aggregation at the likelihood level. These three
approaches target a binary classification problem.  Both
\citet{Kim:MIL:GP:ICML:2010} and \citet{Haubmann:MIL:GP:CVPR:2017}
focus on the case for which the label of the bag corresponds to the
maximum of the (unobserved) individual labels of each
input. \citet{Kim:MIL:GP:ICML:2010} approximate the maximum using a
softmax function computed using a latent GP prior evaluated across the
individual elements of the bag. They use the Laplace approximation for
computing the approximated posterior
\citep{Rasmussen2006}. \citet{Haubmann:MIL:GP:CVPR:2017}, on the other
hand, approximate the maximum using the so called \emph{bag label
  likelihood}, introduced by the authors, which is similar to a
Bernoulli likelihood with soft labels given by a convex combination
between the bag labels and the maximum of the (latent) individual
labels. The latent individual labels in turn follow Bernoulli
likelihoods with parameters given by a GP. The authors provide a
variational bound and include inducing inputs for scalable Bayesian
inference. \citet{Kandemir:MIL:GP:BMVC:2016} follow a similar approach
to \citet{Law:variational:aggregate:neurips:2018} equivalent to
setting all the weights in
\citeauthor{Law:variational:aggregate:neurips:2018}'s model to
one. The sum is then used to modulate the parameter of a Bernoulli
likelihood that models the bag labels. They use a Fully Independent
Training Conditional approximation for the latent GP prior
\citep{Snelson:FITC:2006}.  In contrast to these previous works, we
provide a multi-task learning model for aggregated data that scales to
large datasets and allows for heterogeneous outputs.  At the time of
submission of this paper, the idea of using multi-task learning for
aggregated datasets was simultaneously proposed by
\citet{hamelijnck2019multi} and \citet{tanaka2019spatially}, two
additional models to the one we propose in this paper.  In our
work, we allow heterogenous likelihoods which is different to both
\citet{hamelijnck2019multi} and \citet{tanaka2019spatially}. We also
allow an exact solution to the integration of the latent function
through the kernel in \citet{Smith:binned:2018}, which is different to
\citet{hamelijnck2019multi}. Also, for computational complexity,
inducing inputs are used, another difference from the work in
\citet{tanaka2019spatially}. Other relevant work is described in
Section \ref{sec:related:work}.

For building the multi-task learning model we appeal to the linear model of coregionalisation \citep{Journel:miningBook78, Goovaerts:book97} that has gained popularity in the
multi-task GP literature in recent years
\citep{Bonilla:Multitask:2008, alvarez2012kernels}. We also allow
different likelihood functions \citep{MorenoMunoz:HetMOGP:2018}  and
different input supports per individual task. Moreover, we introduce
inducing inputs at the level of the underlying common set of latent
functions, an idea initially proposed in \citet{Alvarez:NeurIPS:2008}. We then use stochastic variational
inference for GPs \citep{hensman2013gaussian} leading to an approximation similar to the one obtained in
\citet{MorenoMunoz:HetMOGP:2018}. Empirical results show that the multi-task learning approach developed here provides
accurate predictions in different challenging datasets where tasks have different supports. 
\section{Multi-task learning for aggregated data at different scales}
In this section we first define the basic model in the single-task
setting. We then extend the model to the multi-task setting and
finally provide details for the stochastic variational formulation for
approximate Bayesian inference.

\subsection{Change of support using Gaussian processes}\label{sec:change:support}

Change of support has been studied in geostatistics before \citep{Gotway:Incompatible:2002}. In this paper, we use a formulation
similar to \citet{Kyriakidis:AreaToPoint:2004}. We start by defining a stochastic process over the input interval $(x_a, x_b)$ using
\begin{align*}
f(x_a, x_b) = \frac{1}{\Delta_x}\int_{x_a}^{x_b} u(z)dz, 
\end{align*}
where $u(z)$ is a latent stochastic process that we assume follows a Gaussian process with zero mean and covariance $k(z,z')$ and $\Delta_x=|x_b-x_a|$. Dividing by $\Delta_x$ helps to keep the proportionality between the length of the interval and the
area under $u(z)$ in the interval. In other words, the process $f(\cdot)$
is modeled as a density meaning that inputs with widely
differing supports will behave in a similar way. The first two moments for $f(x_a, x_b)$ are
given as $\mathbb{E}[f(x_a, x_b)]= 0$ and $\mathbb{E}[f(x_a, x_b), f(x'_a, x'_b)] = \frac{1}{\Delta_x \Delta_{x'}}\int_{x_a}^{x_b} \int_{x'_a}^{x'_b} \mathbb{E}[u(z)u(z')]dz'dz$. The covariance for $f(x_a, x_b)$
follows as $\cov[f(x_a, x_b), f(x'_a, x'_b)]= \frac{1}{\Delta_x
  \Delta_{x'}}\int_{x_a}^{x_b} \int_{x'_a}^{x'_b} k(z,z')dz'dz$ since
$\mathbb{E}[u(z)]=0$. Let us use $k(x_a, x_b, x'_a, x'_b)$ to refer to
$\cov[f(x_a, x_b), f(x'_a, x'_b)]$. We can now use these mean and covariance functions for representing the Gaussian process prior for
$f(x_a, x_b) \sim \mathcal{GP}(0, k(x_a, x_b, x'_a, x'_b))$. 

For some forms of $k(z,z')$ it is possible to obtain an analytical expression for $k(x_a, x_b, x'_a, x'_b)$. For example, if $k(z,z')$ follows an Exponentiated-Quadratic
(EQ) covariance form, $k(z,z')=\sigma^2\exp\{-\frac{(z-z')^2}{\ell^2}\}$, where $\sigma^2$ is the variance of the kernel and $\ell$ is the length-scale, it can be
shown that $k(x_a, x_b, x'_a, x'_b)$ follows as
\begin{align*}
  k(x_a, x_b, x'_a, x'_b) & = \frac{\sigma^2\ell^2}{2 \Delta_x \Delta_{x'}} \left[h\left(\frac{x_b-x'_a}{\ell}\right) + h\left(\frac{x_a-x'_b}{\ell}\right) - h\left(\frac{x_a-x'_a}{\ell}\right) -
                            h\left(\frac{x_b-x'_b}{\ell}\right) \right], 
\end{align*}
where $h(z) = \sqrt{\pi}z\erf(z) +e^{-z^2}$ with $\erf(z)$, the error function defined as $\erf(z)=\frac{2}{\sqrt{\pi}}\int_0^ze^{-r^2}dr$. Other kernels for $k(z,z')$ also lead
to analytical expressions for $k(x_a, x_b, x'_a, x'_b)$. See for example \citet{Smith:binned:2018}.

So far, we have restricted the exposition to one-dimensional intervals. However, we can define the stochastic process $f$ over a general support $\upsilon$, with measure $|\upsilon|$, using 
\begin{align*}
f(\upsilon) = \frac{1}{|\upsilon|} \int_{\boldz\in v} u(\boldz)d\boldz.
\end{align*}
The support $\upsilon$ generally refers to an area or volume of any shape or size. Following similar assumptions to the ones we used for $f(x_a, x_b)$,
we can build a GP prior to represent $f(\upsilon)$ with covariance $k(\upsilon, \upsilon')$ defined as
$k(\upsilon, \upsilon')= \frac{1}{|\upsilon||\upsilon'|}\int_{\boldz\in\upsilon} \int_{\boldz'\in\upsilon'} k(\boldz,\boldz')d\boldz'
d\boldz$. Let $\boldz\in\mathbb{R}^p$. If the support $\upsilon$ has
a regular shape, e.g. a hyperrectangle, then assumptions on $u(\boldz)$ such as additivity or factorization across input dimensions will lead to kernels that can be expressed
as addition of kernels or product of kernels acting over a single dimension. For example, let $u(\boldz)=\prod_{i=1}^pu_i(z_i)$, where $\boldz=[z_1,\cdots, z_p]^{\top}$,
and a GP over each $u_i(z_i)\sim \mathcal{GP}(0, k(z_i, z'_i))$. If each $k(z_i, z'_i)$ is an EQ kernel, then $k(\upsilon, \upsilon')=\prod_{i=1}^pk(x_{i,a}, x_{i,b}, x'_{i,a}, x'_{i,b})$,
where $(x_{i,a}, x_{i,b})$ and $(x'_{i,a}, x'_{i,b})$ are the
intervals across each input dimension. If the support $\upsilon$ does
not follow a regular shape,
i.e it is a polytope, then we can approximate the double integration by numerical integration inside the support. 

\subsection{Multi-task learning setting}
Our inspiration for multi-task learning is the linear model of
coregionalisation (LMC) \citep{Journel:miningBook78}. This model has connections with other multi-task learning approaches that use kernel methods. For example, \citet{seeger2005semiparametric} and \citet{Bonilla:Multitask:2008} are particular cases of LMC. A detailed review is available in \citet{alvarez2012kernels}. 
In the LMC, each
output (or task in our case) is represented as a linear combination of
a common set of latent Gaussian processes.  Let
$\{u_{q}(\boldz)\}_{q =1}^{Q}$ be a set of $Q$ GPs with zero means and
covariance functions $k_q(\boldz, \boldz')$. Each GP $u_{q}(\boldz)$ is sampled independently and
identically $R_q$ times to produce $\{u^i_{q}(\boldz)\}_{i=1,q=1}^{R_q, Q}$
realizations that are used to represent the outputs.  Let
$\{f_d(\upsilon)\}_{d=1}^D$ be a set of tasks where each task is defined at a different support $\upsilon$. We use the
set of realizations $u^i_{q}(\boldz)$ to represent each task $f_d(\upsilon)$ as
\begin{align}
  f_d(\upsilon) & = \sum_{q=1}^Q\sum_{i=1}^{R_q}\frac{a^i_{d,q}}{|\upsilon|} \int_{\boldz\in v} u_q^i(\boldz)d\boldz,\label{eq:lmc:area}
\end{align}
where the coefficients $a^i_{d,q}$ weight the contribution of each
integral term to $f_d(\upsilon)$.  Since $\cov[u_q^i(\boldz), u_{q'}^{i'}(\boldz')]= k_q(\boldz, \boldz')\delta_{q,q'}\delta_{i,i'}$, with
$\delta_{\alpha,\beta}$ the Kronecker delta between $\alpha$ and $\beta$, the
cross-covariance $k_{f_d,f_{d'}}(\upsilon, \upsilon')$ between
$f_d(\upsilon)$ and $f_{d'}(\upsilon')$ is then given as
\begin{align*}
  k_{f_d,f_{d'}}(\upsilon, \upsilon') =  \sum_{q=1}^Q \frac{b^q_{d,d'}}{|\upsilon||\upsilon'|}\int_{\boldz\in \upsilon} \int_{\boldz'\in \upsilon'}
  k_q(\boldz,\boldz')d\boldz' d\boldz,
\end{align*}  
where $b^q_{d,d'} = \sum_{i=1}^{R_q} a^i_{d,q} a^i_{d',q}$. Following the discussion in Section \ref{sec:change:support}, the double integral
can be solved analytically for some options of $\upsilon$, $\upsilon'$ and $k_q(\boldz,\boldz')$. Generally, a numerical approximation can be
obtained.

It is also worth mentioning at this point that the model does not require all the tasks to be defined at the area
level. Some of the tasks could also be defined at the point level. Say for example that $f_d$ is defined at the
support level $\upsilon$, $f_{d}(\upsilon)$, whereas $f_{d'}$ is defined at the point level, say
$\boldx\in \mathbb{R}^p$, $f_{d'}(\boldx)$. In this case, $f_{d'}(\boldx) = \sum_{q=1}^Q\sum_{i=1}^{R_q}a^i_{d',q}
u^i_q(\boldx)$. We can still compute the cross-covariance between $f_{d}(\upsilon)$ and
$f_{d'}(\boldx)$, $k_{f_d,f_{d'}}(\upsilon, \boldx)$, leading to, $k_{f_d,f_{d'}}(\upsilon, \boldx) =  \sum_{q=1}^Q
\frac{b^q_{d,d'}}{|\upsilon|}\int_{\boldz\in v}
k_q(\boldz,\boldx)d\boldz$. For the case $Q=1$ and $p=1$
(i.e. dimensionality of the input space), this is, $z,z',x\in\mathbb{R}$, $\upsilon=(x_a, x_b)$ 
and an EQ kernel for $k(z,z')$, we get
$k_{f_d,f_{d'}}(\upsilon, x) =  \frac{b_{d,d'}}{\Delta_x}\int_{x_a}^{x_b} k(z,x)dz = \frac{b_{d,d'}\ell}{2\Delta_x}\left[\erf\left(\frac{x_b-x}{\ell}\right)+\erf\left(\frac{x-x_a}{\ell}\right)\right]$ (we used $\sigma^2=1$ to avoid an overparameterization for the variance). Again, if $\upsilon$ does not have a regular shape, 
we can approximate the integral numerically.

Let us define the vector-valued function $\boldf(\upsilon) = [f_1(\upsilon), \cdots, f_D(\upsilon)]^\top$.
A GP prior over $\boldf(\upsilon)$ can use the kernel defined above so that
\begin{align*}
\boldf(\upsilon)\sim \mathcal{GP}\left(\mathbf{0}, \sum_{q=1}^Q\frac{1}{|\upsilon||\upsilon'|}\mathbf{B}_q \int_{\boldz\in v} \int_{\boldz'\in v'}
  k_q(\boldz,\boldz')d\boldz' d\boldz\right),
\end{align*}
where each $\mathbf{B}_q\in\mathbb{R}^{D\times D}$ is known as a
coregionalisation matrix. The scalar term $\int_{\boldz\in v} \int_{\boldz'\in v'}
  k_q(\boldz,\boldz')d\boldz' d\boldz$ modulates $\mathbf{B}_q$ as a
  function of $\upsilon$ and $\upsilon'$.

The prior above can be used for modulating the parameters of
likelihood functions that model the observed data. The most simple
case corresponds to the multi-task regression problem that can be
modeled using a multivariate Gaussian distribution. Let
$\boldy(\upsilon) = [y_1(\upsilon), \cdots, y_D(\upsilon)]^\top$ be a
random vector modeling the observed data as a function of $\upsilon$. In the multi-task regression
problem
$\boldy(\upsilon)\sim \mathcal{N}(\bm{\mu}(\upsilon), \bm{\Sigma})$,
where
$\bm{\mu}(\upsilon) = [\mu_1(\upsilon), \cdots, \mu_D(\upsilon)]^\top$
is the mean vector and $\bm{\Sigma}$ is a diagonal matrix with entries
$\{\sigma_{y_d}^2 \}_{d=1}^D$. We can use the GP prior
$\boldf(\upsilon)$ as the prior over the mean vector
$\bm{\mu}(\upsilon) \sim \boldf(\upsilon)$. Since both the likelihood
and the prior are Gaussian, both the marginal distribution for
$\boldy(\upsilon)$ and the posterior distribution of
$\boldf(\upsilon)$ given $\boldy(\upsilon)$ can be computed
analytically. For example, the marginal distribution for
$\boldy(\upsilon)$ is given as
$\boldy(\upsilon) \sim \mathcal{N}(\mathbf{0},
\sum_{q=1}^Q\frac{1}{|\upsilon||\upsilon'|}\mathbf{B}_q \int_{\boldz\in v} \int_{\boldz'\in v'}
k_q(\boldz,\boldz')d\boldz' d\boldz+\bm{\Sigma})$.
\citet{MorenoMunoz:HetMOGP:2018} introduced the idea of allowing each
task to have a different likelihood function and modulated the
parameters of that likelihood function using one or more elements in
the vector-valued GP prior. For that general case, the marginal likelihood
and the posterior distribution cannot be computed in closed form.
\subsection{Stochastic variational inference}
Let $\mathcal{D} = \{\bm{\Upsilon}, \mathbf{y}\}$ be a dataset of
multiple tasks with potentially different supports per task, where
$\bm{\Upsilon} = \{\bm{\upsilon}_d\}_{d=1}^D$, with
$\bm{\upsilon}_d=[\upsilon_{d,1},\cdots, \upsilon_{d,N_d}]^{\top}$,
and $\mathbf{y}=[\boldy_1, \cdots, \boldy_D]^{\top}$, with
$\boldy_d=[y_{d,1}, \cdots, y_{d,N_d}]^{\top}$ and
$y_{d, j}=y_d(\upsilon_{d,j})$. Notice that $\mathbf{y}$ without $\upsilon$ refers to the output vector for the dataset.
We are interested in computing the
posterior distribution
$p(\boldf|\boldy)= p(\boldy|\boldf)p(\boldf)/p(\boldy)$, where
$\mathbf{f}=[\boldf_1, \cdots, \boldf_D]^{\top}$, with
$\boldf_d=[f_{d,1}, \cdots, f_{d,N_d}]^{\top}$ and
$f_{d, j}=f_d(\upsilon_{d,j})$. In this paper, we will use stochastic
variational inference to compute a deterministic approximation of the
posterior distribution $p(\boldf|\boldy)\approx q(\boldf)$, by means
of the the well known idea of \textit{inducing variables}.
In contrast to the use of SVI for traditional Gaussian
processes, where the inducing variables are defined at the level of
the process $\boldf$, we follow \citet{alvarez2010efficient} and 
  \citet{MorenoMunoz:HetMOGP:2018}, and define the inducing variables at the
level of the latent processes $u_q(\boldz)$. For simplicity in the
notation, we assume $R_q=1$. Let $\boldu=\{\boldu_q\}_{q=1}^{Q}$ be
the set of inducing variables, where
$\boldu_q = [u_q(\boldz_1),\cdots, u_q(\boldz_{M})]^{\top}$, with
$\boldZ=\{\boldz_m\}_{m=1}^M$ the inducing inputs. Notice also that we
have used a common set of inducing inputs $\boldZ$ for all latent
functions but we can easily define a set $\boldZ_q$ per inducing
vector $\boldu_q$.

For the multi-task regression case, it is possible to compute an
analytical expression for the Gaussian posterior distribution over the
inducing variables $\boldu$, $q(\boldu)$, following a similar approach
to \citet{alvarez2010efficient}. However, such approximation is only
valid for the case in which the likelihood model $p(\boldy|\boldf)$ is
Gaussian and the variational bound obtained is not amenable for
stochastic optimisation. An alternative for finding $q(\boldu)$ also
establishes a lower-bound for the log-marginal likelihood
$\log p(\boldy)$, but uses numerical optimisation for maximising the
bound with respect to the mean parameters, $\bm{\mu}$, and the
covariance parameters, $\mathbf{S}$, for the Gaussian distribution
$q(\boldu)\sim\mathcal{N}(\bm{\mu},\mathbf{S})$ \citep{MorenoMunoz:HetMOGP:2018}. Such
numerical procedure can be used for any likelihood model
$p(\boldy|\boldf)$ and the optimisation can be performed using
mini-batches. We follow this approach.

\paragraph{Lower-bound} The lower bound for the log-marginal likelihood follows as
\begin{align*}
  \log p(\boldy) \ge \int \int q(\boldf, \boldu) \log\frac{p(\boldy|\boldf)p(\boldf|\boldu)p(\boldu)}{q(\boldf, \boldu)}d\boldf  d\boldu =\mathcal{L},
  \end{align*}
where $q(\boldf, \boldu)=p(\boldf|\boldu)q(\boldu)$, $p(\boldf|\boldu)\sim \mathcal{N}(\boldK_{\boldf\boldu}\boldK^{-1}_{\boldu\boldu}
\boldu,\boldK_{\boldf\boldf} -\boldK_{\boldf\boldu}\boldK^{-1}_{\boldu\boldu}\boldK_{\boldf\boldu}^{\top})$, and
$p(\boldu) \sim\mathcal{N}(\mathbf{0}, \boldK_{\boldu\boldu})$  is the prior over the inducing variables. Here $\boldK_{\boldf\boldu}$ is a
blockwise matrix with matrices $\boldK_{\boldf_d, \boldu_q}$. In turn each of these matrices have entries given by
$k_{f_d, u_q}(\upsilon, \boldz')= \frac{a_{d,q}}{|\upsilon|}\int_{\boldz\in\upsilon} k_q(\boldz, \boldz')d\boldz$. Similarly, $\boldK_{\boldu\boldu}$ is a
block-diagonal matrix with blocks given by $\boldK_{q}$ with entries computed using $k_{q}(\boldz, \boldz')$. The optimal $q(\boldu)$ is chosen
by numerically maximizing $\mathcal{L}$ with respect to the parameters $\bm{\mu}$ and $\mathbf{S}$. To ensure a valid covariance matrix
$\mathbf{S}$ we optimise the Cholesky factor $\mathbf{L}$ for $\mathbf{S}=\mathbf{L}\mathbf{L}^{\top}$. See Appendix \ref{sec:appendix:inference:more} for
more details on the lower bound. The computational complexity is similar to the one for the model in \citet{MorenoMunoz:HetMOGP:2018}, $\mathcal{O}(QM^3+JNQM^2)$, where $J$ depends on the types of likelihoods used for the different tasks.
For example, if we model all the outputs using Gaussian likelihoods, then $J=D$. For details, see
\citet{MorenoMunoz:HetMOGP:2018}.

\paragraph{Hyperparameter learning} When using the multi-task learning
method, we need to optimise the hyperparameters associated with the
LMC, these are: the coregionalisation matrices $\mathbf{B}_q$, the
hyperparameters of the kernels $k_q(\boldz, \boldz')$, and any other
hyperparameter associated to the likelihood functions
$p(\boldy|\boldf)$ that has not been considered as a member of the
latent vector $\boldf(\upsilon)$. Hyperparameter optimisation is done
using the lower bound $\mathcal{L}$ as the objective function. First
$\mathcal{L}$ is maximised with respect to the variational distribution
$q(\boldu)$ and then with respect to the hyperparameters. The
two-steps are repeated one after the other until reaching
convergence. Such style of optimisation is known as variational EM
(Expectation-Maximization) when using the full dataset \citep{beal2003variational} or stochastic version, when employing mini-batches \citep{hoffman2013stochastic}. In the Expectation step we compute a variational posterior distribution and in the Maximization step we use a variational lower bound to find point estimates of any
hyperparameters. For optimising the hyperparameters in $\boldB_q$, we
also use a Cholesky decomposition for each matrix to ensure positive
definiteness. So instead of optimising $\boldB_q$ directly, we
optimise $\mathbf{L}_q$, where
$\boldB_q= \mathbf{L}_q \mathbf{L}_q^{\top}$. For the experimental section, we use the EQ kernel for $k_q(\boldz, \boldz)$, so we fix
the variance for $k_q(\boldz, \boldz)$ to one (the variance per output is already contained in the matrices $\boldB_q$) and optimise the length-scales $\ell_q$. 
\paragraph{Predictive distribution} Given
a new set of test inputs $\bm{\Upsilon}_*$, the predictive distribution for $p(\boldy_*|\boldy, \bm{\Upsilon}_*)$ is computed using
$p(\boldy_*|\boldy, \bm{\Upsilon}_*)=\int_{\boldf_*}p(\boldy_*|\boldf_*)q(\boldf_*)d\boldf_*$, where $\boldy_*$ and $\boldf_*$ refer to the vector-valued functions
$\boldy$ and $\boldf$ evaluated at $\bm{\Upsilon}_*$. Notice that
$q(\boldf_*)\approx p(\boldf_*|\boldy)$. Even though $\boldy$ does not appear explicitly in the
expression for $q(\boldf_*)$, it has been used to compute the posterior for $q(\boldu)$ through the optimisation of $\mathcal{L}$ where $\boldy$ is explicitly taken
into account. We are usually interested in the mean prediction $\mathbb{E}[\boldy_*]$ and the predictive variance $\operatorname{var}[\boldy_*]$.
Both can be computed by exchanging integrals in the double integration over $\boldy_*$ and $\boldf_*$. See Appendix \ref{sec:appendix:inference:more} for
more details on this. 
\section{Related work}\label{sec:related:work} 
Machine learning methods for different forms of aggregated datasets
are also known under the names of \emph{multiple instance learning},
\emph{learning from label proportions} or \emph{weakly supervised
  learning on aggregate outputs}
\citep{Kuck:Individuals:ti:Groups:2005,
  Musicant:Sup:learning:aggregate:2007, Quadrianto:JMLR:2009,
  Patrini:NoLabelNoCry:2014,
  Kotzias:Individual:Group:Deep:Features:2015,
  Bhowmik:Aggregated:AISTATS:2015}. \citet{Law:variational:aggregate:neurips:2018}
provided a summary of these different approaches. Typically these
methods start with the following setting: each instance in the dataset
is in the form of a set of inputs for which there is only
one corresponding output (e.g. the proportion of class labels, an average
or a sample statistic). The prediction problem usually consists then
in predicting the individual outputs for the individual inputs in the set.  The
setting we present in this paper is slightly different in the sense
that, in general, for each instance, the input corresponds to a
support of any shape and size and the output corresponds to a
vector-valued output. Moreover, each task can have its own support.
Similarly, while most of these ML approaches have been developed for
either regression or classification, our model is built on top of
\citet{MorenoMunoz:HetMOGP:2018}, allowing each task to have a
potentially different likelihood.

As mentioned in the introduction, Gaussian processes have also been used for multiple instance learning
or aggregated data \citep{Kim:MIL:GP:ICML:2010, Kandemir:MIL:GP:BMVC:2016,
  Haubmann:MIL:GP:CVPR:2017, Smith:binned:2018, Law:variational:aggregate:neurips:2018, tanaka2019refining,
  hamelijnck2019multi, tanaka2019spatially}.
Compared to most of these previous approaches, our model goes beyond the
single task problem and allows learning multiple tasks
simultaneously. Each task can have its own support at training and
test time. Compared to other multi-task approaches,
we allow for heterogeneous outputs. Although our model was
formulated for a continuous support $\boldx\in \upsilon_{d,j}$, we can also define it in terms of a finite set of (previously defined) inputs in the support, e.g. a set
$\{\boldx_{d, j, 1}, \cdots, \boldx_{d, j, K_{d,j}}\}\in \upsilon_{d,j}$ which is more akin to the bag formulations in these previous works. This would require changing the
integral
$\frac{1}{|\upsilon_{d,j}|}\int_{\boldz\in\upsilon_{d,j}}u_q^i(\boldz)d\boldz$
in \eqref{eq:lmc:area} for the sum $\frac{1}{K_{d,j}}\sum_{\forall
  \boldx\in \upsilon_{d,j}}u_q^i(\boldx_{d,j,k})$.

In geostatistics, a similar problem has been studied under the names of \emph{downscaling} or \emph{spatial disaggregation}
\citep{zhang2014scale}, particularly using different forms of \emph{kriging} \citep{Goovaerts:book97}. It
is also closely related to the problem of \emph{change of support} described with detail in \citet{Gotway:Incompatible:2002}.
Block-to-point kriging (or area-to-point kriging if the support is defined in a surface) is a common method for downscaling,
this is, provide predictions at the point level provided data at the
block level \citep{Kyriakidis:AreaToPoint:2004,
  GoovaertsSupport2010}. We extend the approach introduced in
\citet{Kyriakidis:AreaToPoint:2004} later revisited by
\citet{GoovaertsSupport2010} for count data, to the multi-task
setting, including also a stochastic variational EM algorithm for
scalable inference.

If we consider the high-resolution outputs as high-fidelity outputs
and low-resolution outputs as low-fidelity outputs, our work also
falls under the umbrella of \emph{multi-fidelity models} where co-kriging
using the linear model of coregionalisation has also been used as
an alternative \citep{Peherstorfer:Multifidelity:2018, fernndezgodino2016review}.
\section{Experiments}
In this section, we apply the multi-task learning model for prediction
in three different datasets: a synthetic example for two tasks that each have a Poisson likelihood, a two-dimensional input dataset of fertility rates aggregated by year of conception and ages in Canada, and an
air-pollution sensor network where one task corresponds to a
high-accuracy, low-frequency particulate matter sensor and another task
corresponds to a low-cost, low-accuracy, high resolution sensor. 
In these examples, we use $k$-means clustering over the input data, with $k=M$, to initialise the values of the inducing inputs, $\mathbf{Z}$, which are also kept fixed during optimisation. We assume the inducing inputs are points, but they could be defined as intervals or supports. For standard optimisation we used the LBFGS-B algorithm and when SVI was needed, the Adam optimiser, included in \textit{climin}
library, was used for the optimisation of the variational distribution (variational E-step) and the hyperparameters (variational M-step). 
The implementation is based on the GPy framework and is available on Github: \url{https://github.com/frb-yousefi/aggregated-multitask-gp}.

\paragraph{Synthetic data}
\input{synthetic}
\paragraph{Fertility rates from a Canadian census}
\input{fertility}
\paragraph{Air pollution monitoring network}
\input{airpollution}

\section{Conclusion}

In this paper, we have introduced a powerful framework
for working with aggregated datasets that allows the user to combine
observations from disparate data types, with varied support. This
allows us to produce both finely resolved and accurate predictions by
using the accuracy of low-resolution data and the fidelity of
high-resolution side-information. We chose our inducing points to lie
in the latent space, a distinction which allows us to estimate
multiple tasks with different likelihoods. SVI and variational-EM with
mini-batches make the framework scalable and tractable for potentially
very large problems. A potential extension would be to consider how
the ``mixing'' achieved through coregionalisation could vary across the
domain by extending, for example, the Gaussian Process Regression Network
model \citep{wilson2011gaussian} to be able to deal with aggregated data. Such model would allow latent
functions of different lengthscales to be relevant at different
locations in the domain. In summary, this framework provides a vital
toolkit, allowing a mixture of likelihoods, kernels and tasks and
paves the way to the analysis of a very common and widely used data
structure - that of values over a variety of supports on the domain.

\section{Acknowledgement}
MTS and MAA have been financed by the Engineering and Physical Research Council (EPSRC) Research Project EP/N014162/1. MAA has also been financed by the EPSRC Research Project EP/R034303/1.

\small
\bibliographystyle{plainnat} 
\bibliography{refslowhigh}

\newpage

\appendix

\section{Supplemental material}

\subsection{Additional details on SVI}\label{sec:appendix:inference:more}

\paragraph{Lower-bound} It can be shown
\citep{MorenoMunoz:HetMOGP:2018} that the bound is given as
\begin{align*}
  \mathcal{L} & = \sum_{d=1}^D\sum_{j=1}^{N_d} \mathbb{E}\left[\log p(y_d(\upsilon_{d,j})|f_{d}(\upsilon_{d,j}))\right] - \operatorname{KL}(q(\boldu)\|p(\boldu)),
\end{align*}  
where the expected value is taken with respect to the $q(\boldf)=\int
q(\boldf, \boldu)d\boldu$ distribution, which is a Gaussian distribution with mean
$\boldK_{\boldf\boldu}\boldK^{-1}_{\boldu\boldu}\bm{\mu}$ and covariance $\boldK_{\boldf \boldf} + \boldK_{\boldf\boldu}\boldK^{-1}_{\boldu\boldu} (\mathbf{S}-\boldK_{\boldu\boldu}) \boldK^{-1}_{\boldu\boldu}\boldK_{\boldf\boldu}^{\top}$. For Gaussian likelihoods,
\begin{align*}
  p(y_d(\upsilon_{d,j})|f_{d}(\upsilon_{d,j}))=\mathcal{N}(y_d(\upsilon_{d,j})|f_d(\upsilon_{d,j}), \sigma^2_{y_d}),
\end{align*}
we can compute the expected value in the bound in closed form. For other likelihoods, we can use numerical integration to approximate it such as Gaussian-Hermite quadratures as in \citet{hensman2015scalable} and \citet{Saul2016chained}. Instead of using the whole batch of data $N=\sum_{d=1}^DN_d$, we can use mini-batches to estimate the gradient.

\paragraph{Predictive distribution} As we mentioned before, we are usually interested in the mean prediction $\mathbb{E}[\boldy_*]$ and
the predictive variance $\operatorname{var}[\boldy_*]$. Both can be computed by exchanging integrals in the double integration over $\boldy_*$ and $\boldf_*$ For example,
$\mathbb{E}[\boldy_*]=\int_{\boldy_*}\boldy_*p(\boldy_*|\boldy, \bm{\Upsilon}_*)d\boldy_*$$=\int_{\boldf_*}\int_{\boldy_*}\boldy_*
p(\boldy_*|\boldf_*) d\boldy_*q(\boldf_*)d\boldf_*$. The inner
integral in $\mathbb{E}[\boldy_*]$ is computed with the conditional distribution $p(\boldy_*|\boldf_*)$ and its form depend on the likelihood
term per task. The outer integral can be approximated using numerical integration or Monte-Carlo sampling. A similar procedure can be followed to compute $\operatorname{var}[\boldy_*]$.

\subsection{SNLP for the fertility dataset}\label{appendix:sec:fertility:snlp}

\begin{figure}[H]
	\centering
		\includegraphics[width=0.45\textwidth]{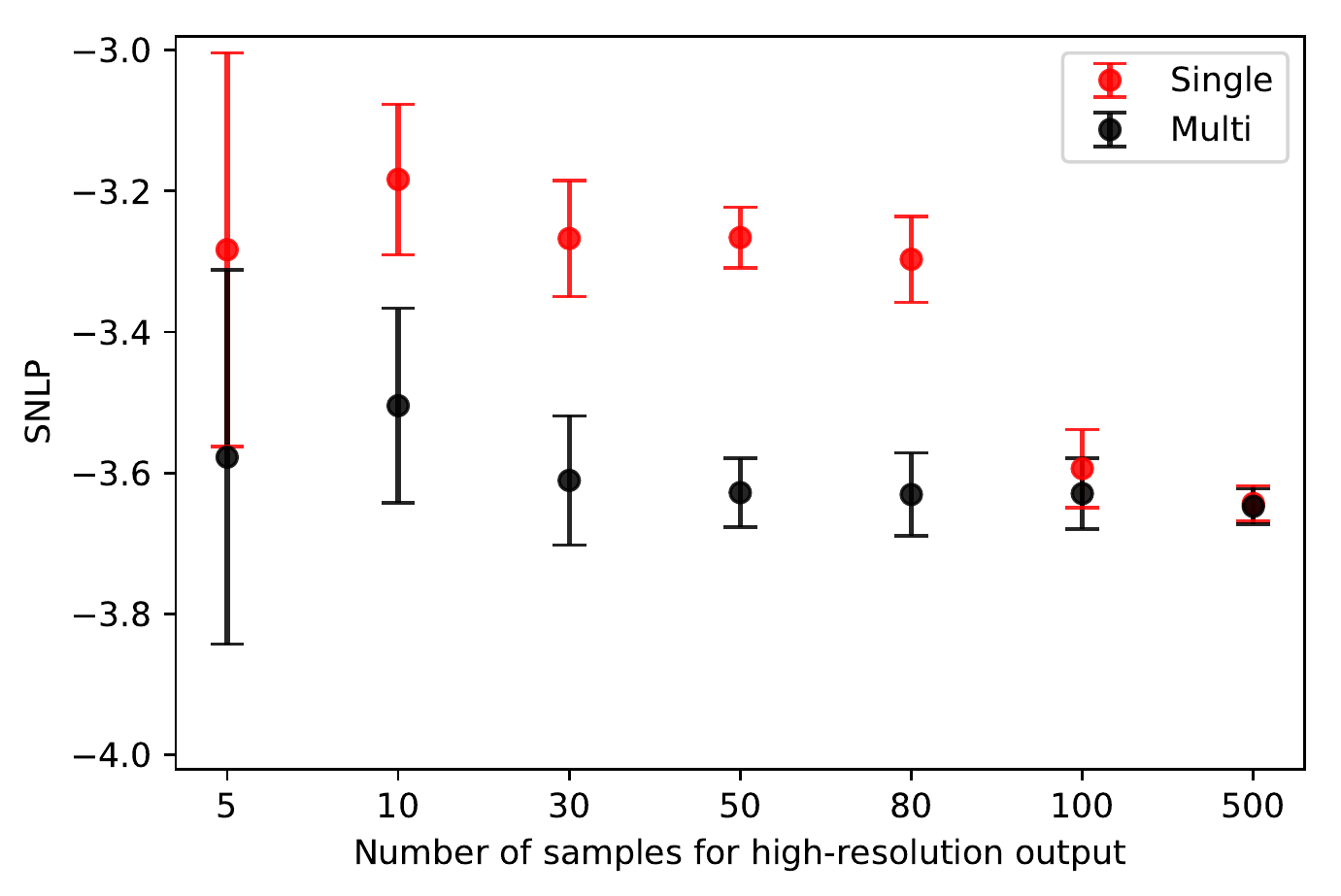}
		\includegraphics[width=0.45\textwidth]{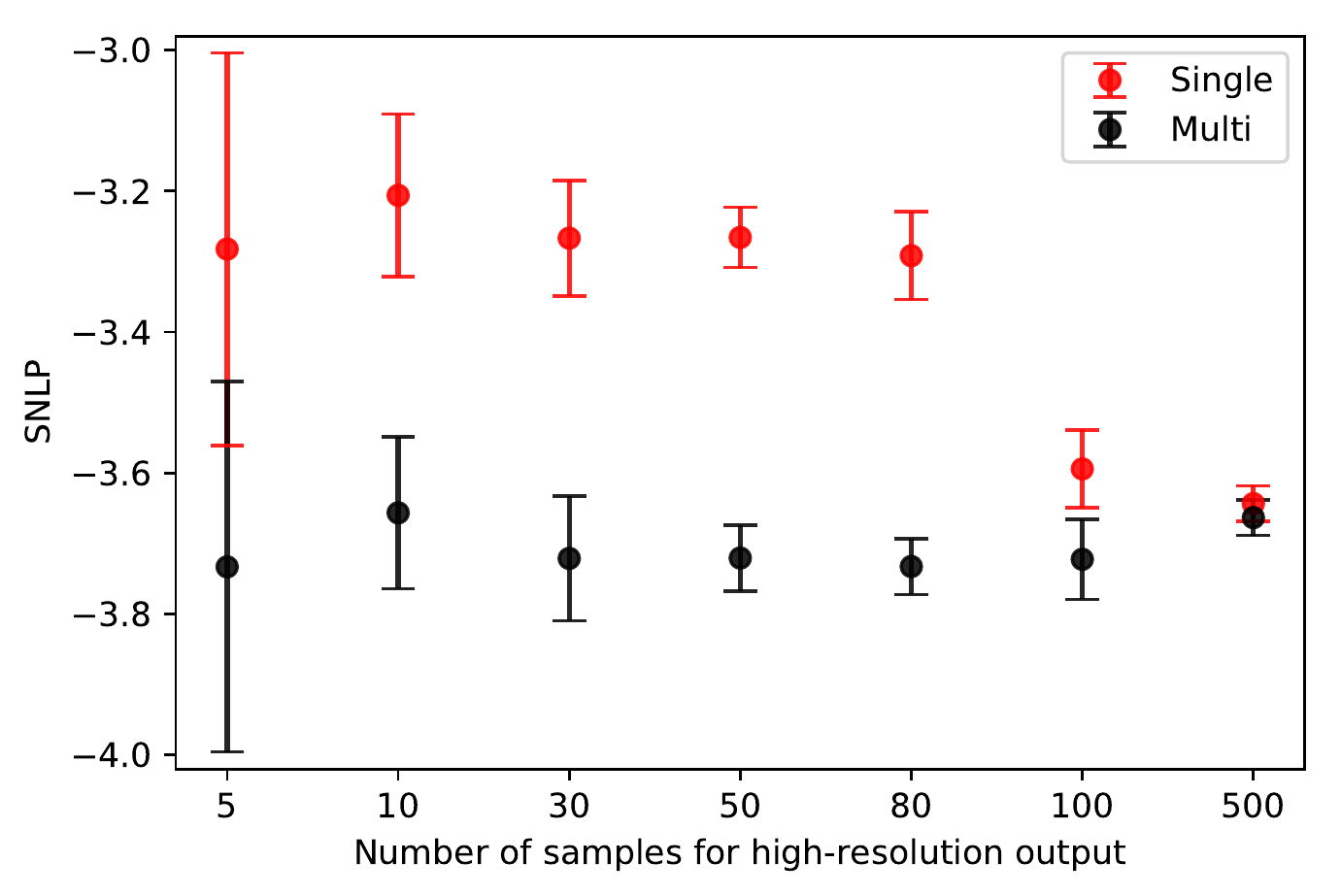}
		\caption{SNLP for $5\times 5$ and
                  $2\times 2$ aggregated data}\label{fig:snlp:fertility}
\end{figure}

Figure \ref{fig:snlp:fertility} shows the results in terms of SNLP for the Fertility
dataset. We can notice a similar pattern to the one observed for the SMSE in Figure \ref{fig:fertility:smse}.

\subsection{Comparing the model with different baselines for the fertility dataset} \label{appendix:sec:fertility:baselines}
\begin{figure}[H]
	\centering
	\includegraphics[width=0.45\textwidth]{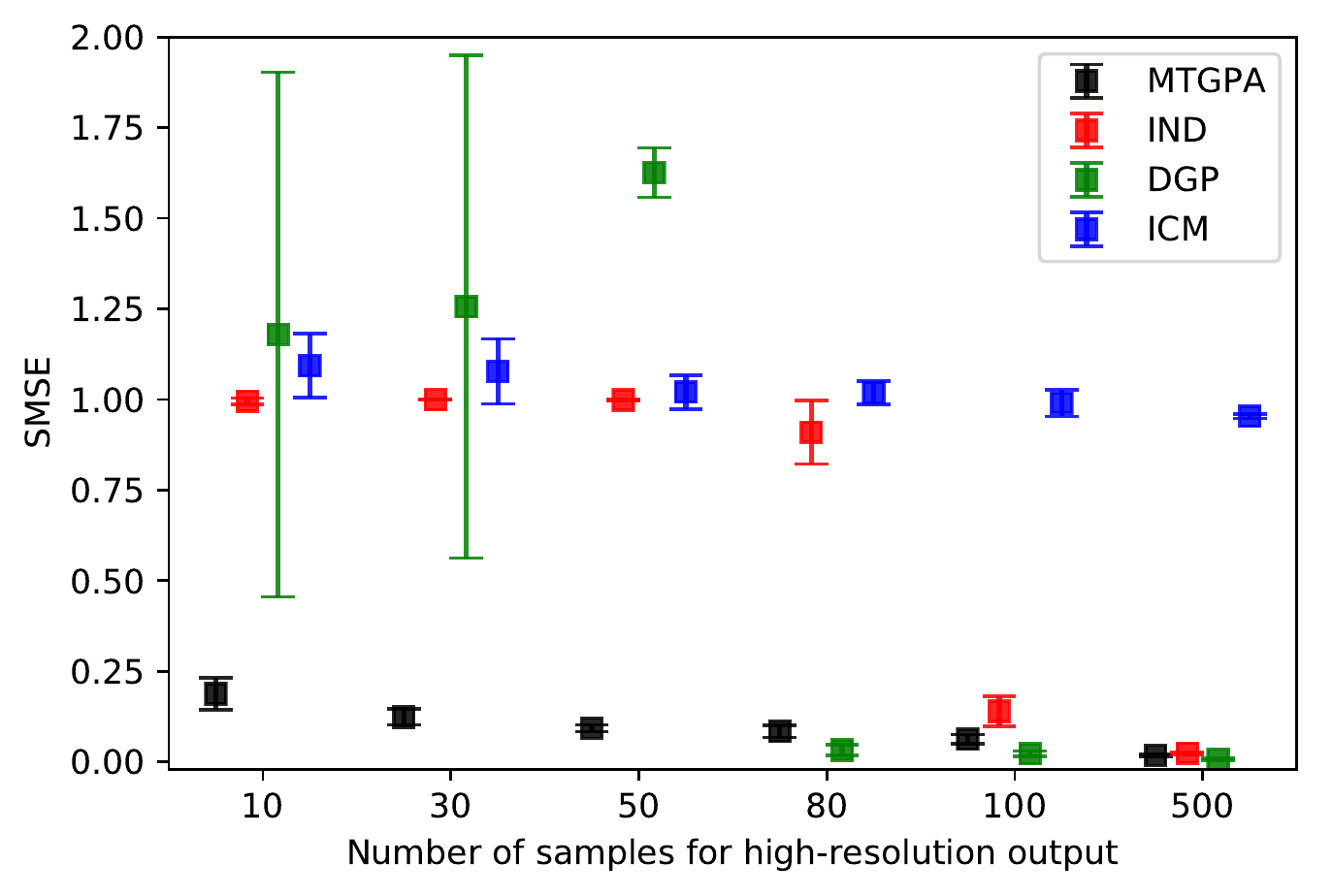}
	\includegraphics[width=0.45\textwidth]{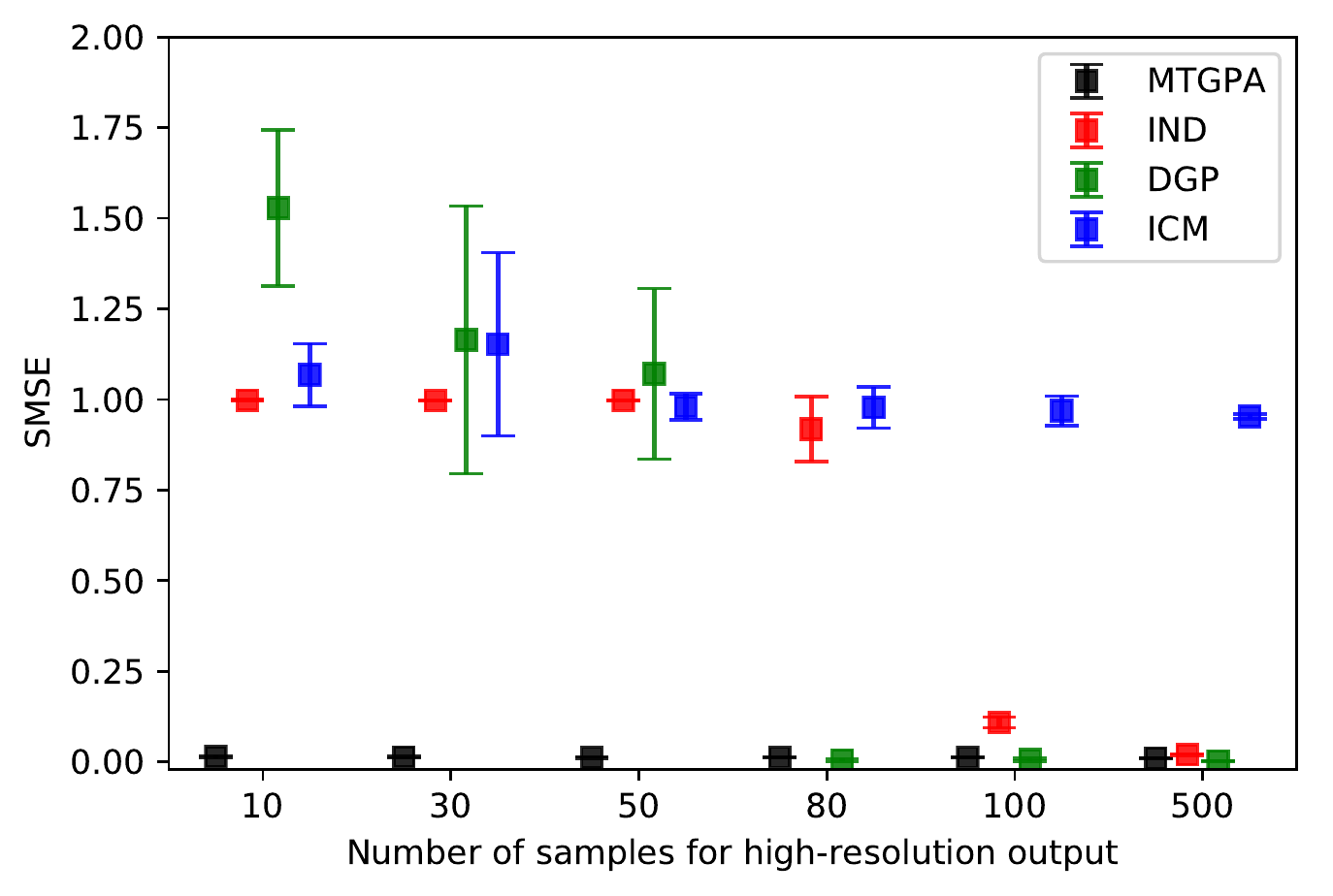}		
	\includegraphics[width=0.45\textwidth]{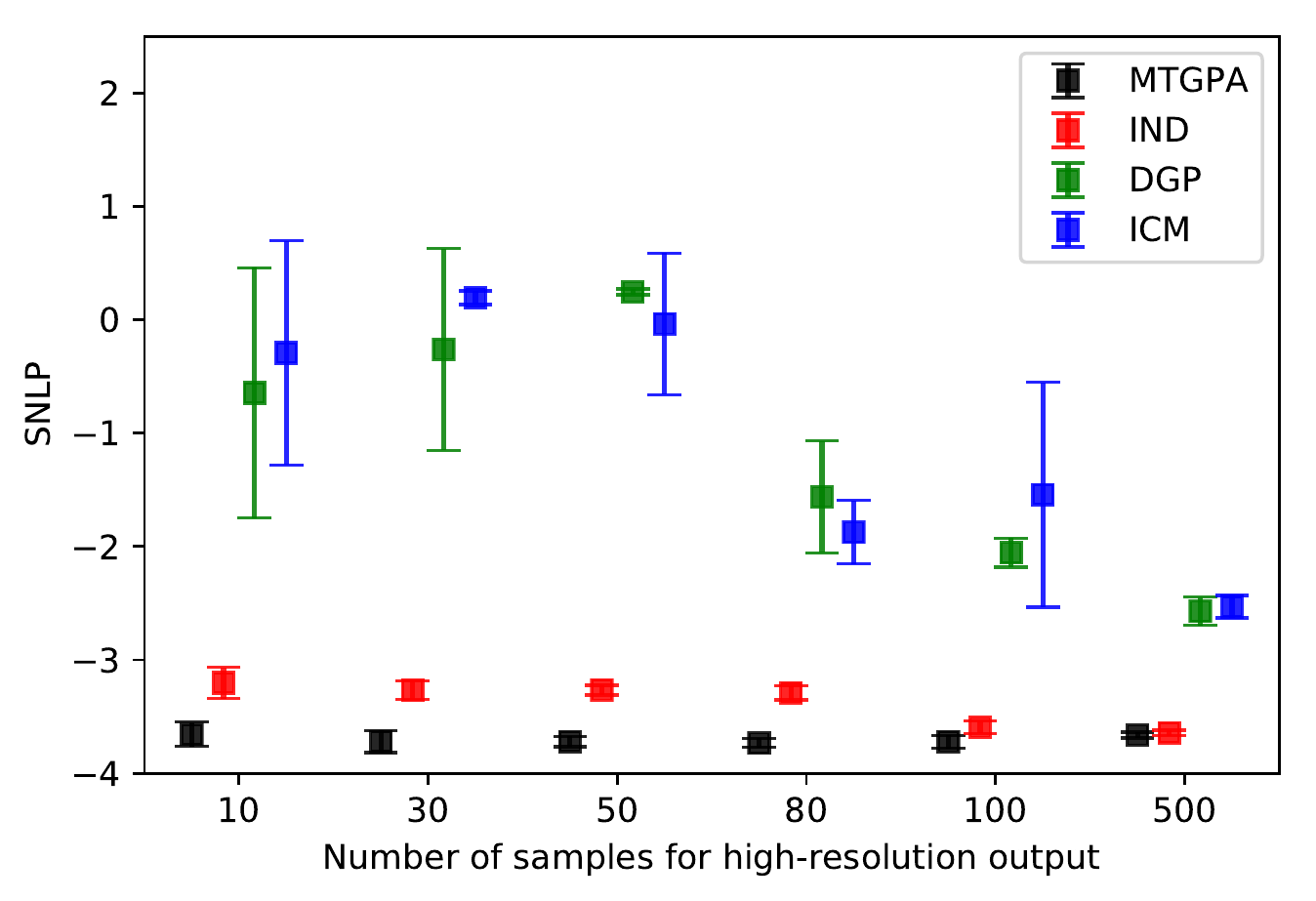}
	\includegraphics[width=0.45\textwidth]{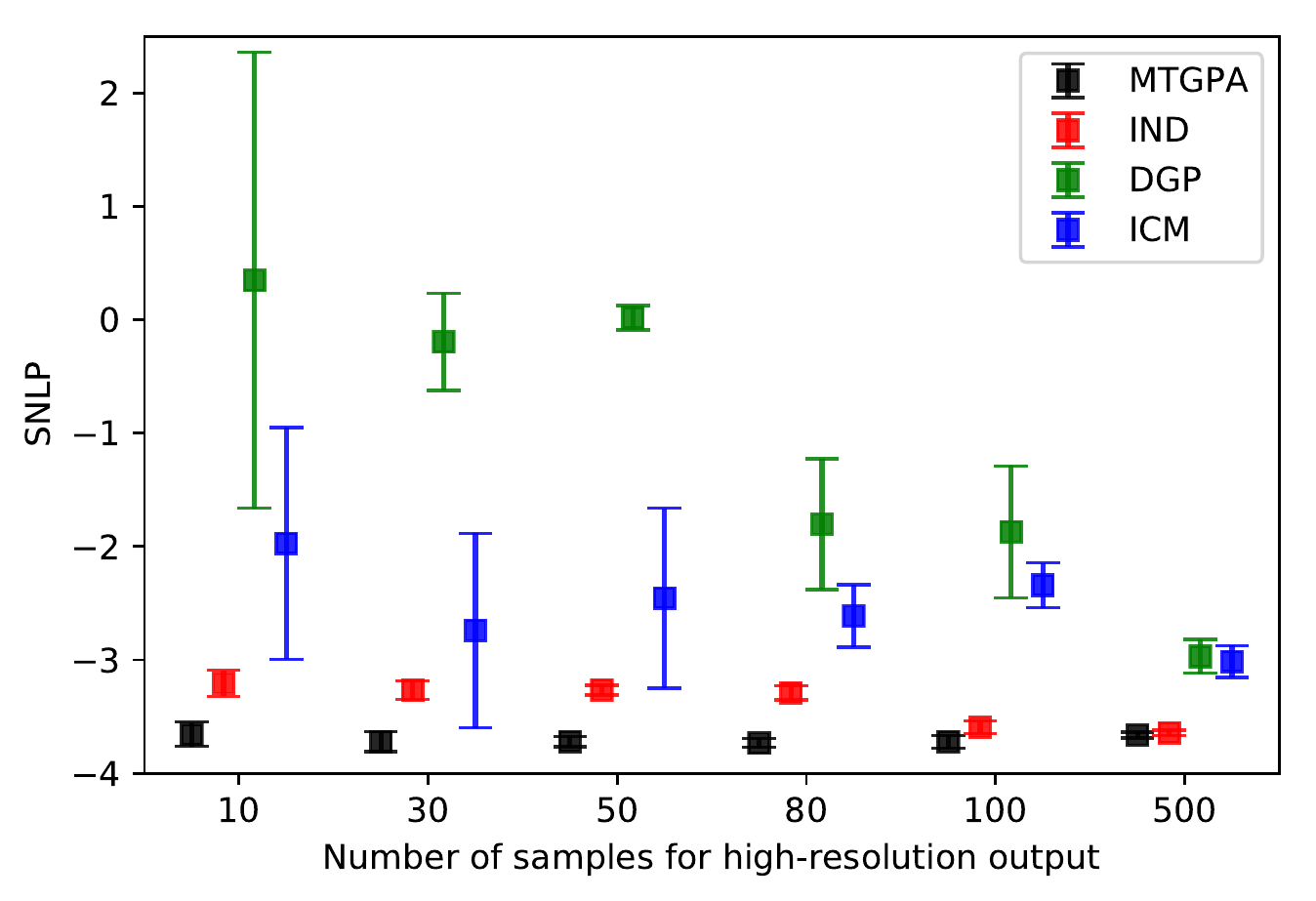}		
	\caption{SMSE and SNLP plots for the fertility dataset for $5\times 5$ (left panel) and $2\times 2$ (right panel) aggregated data for different baselines, MTGPA , Independent GP (IND), DGP and ICM.}
	\label{fig:smse:baselinefertility}
\end{figure}

In Figure \ref{fig:smse:baselinefertility} different baselines are compared to the proposed method. Dependent GPs (DGP) \citep{Boyle2004DGP} and Intrinsic Co-regionalisation Model or Multi-task GPs (ICM) \citep{Bonilla:Multitask:2008} use the centroid of the area as input. MTGPA (proposed method) performs better or similar to baselines as we increase the number of training points for the high-resolution output.

\subsection{Experimental results considering more tasks for the fertility dataset}\label{appendix:sec:fertility:hetgp:toy}
\begin{figure}[H]
	\centering
	\includegraphics[width=0.45\textwidth]{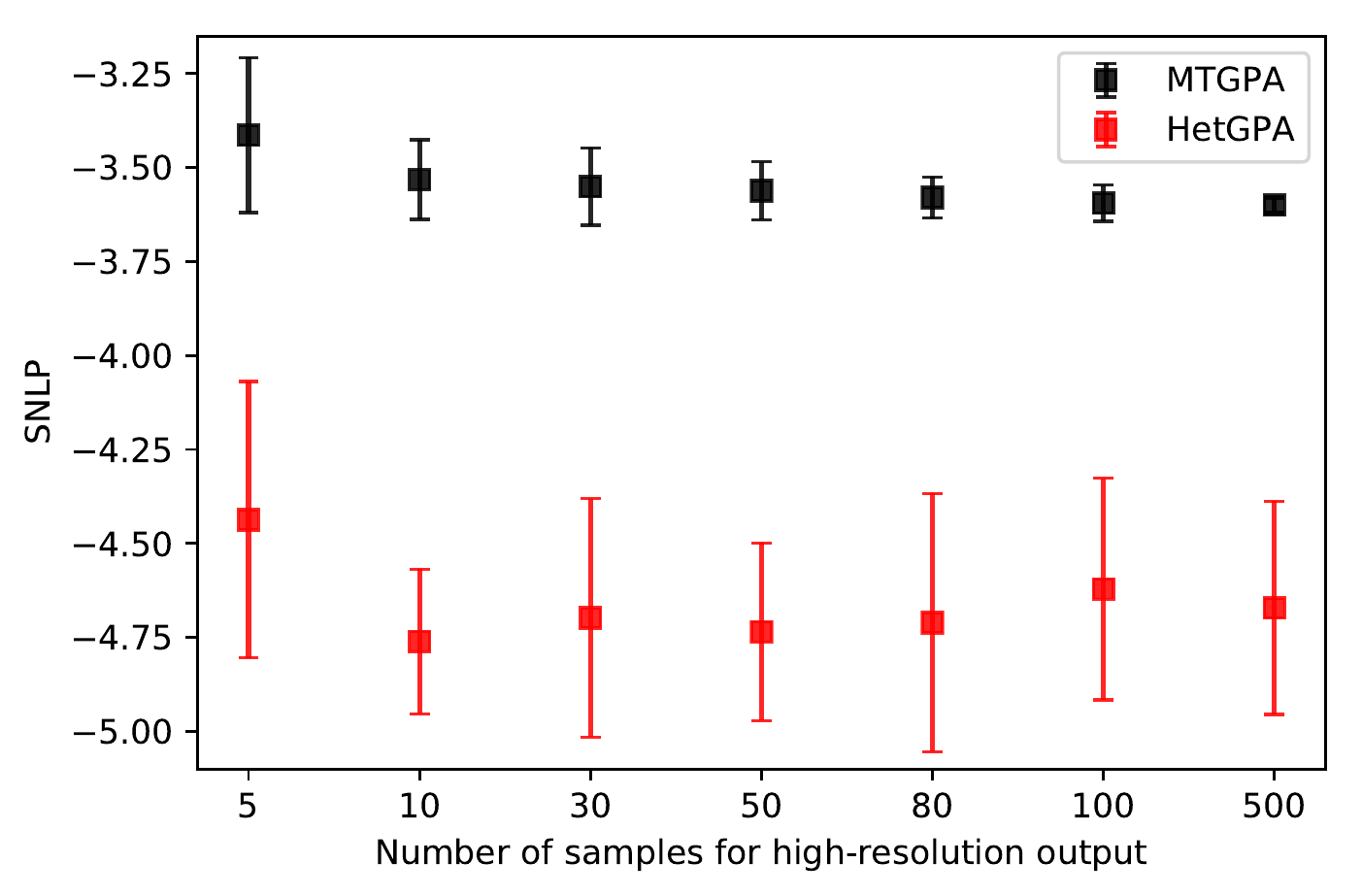}
	\includegraphics[width=0.45\textwidth]{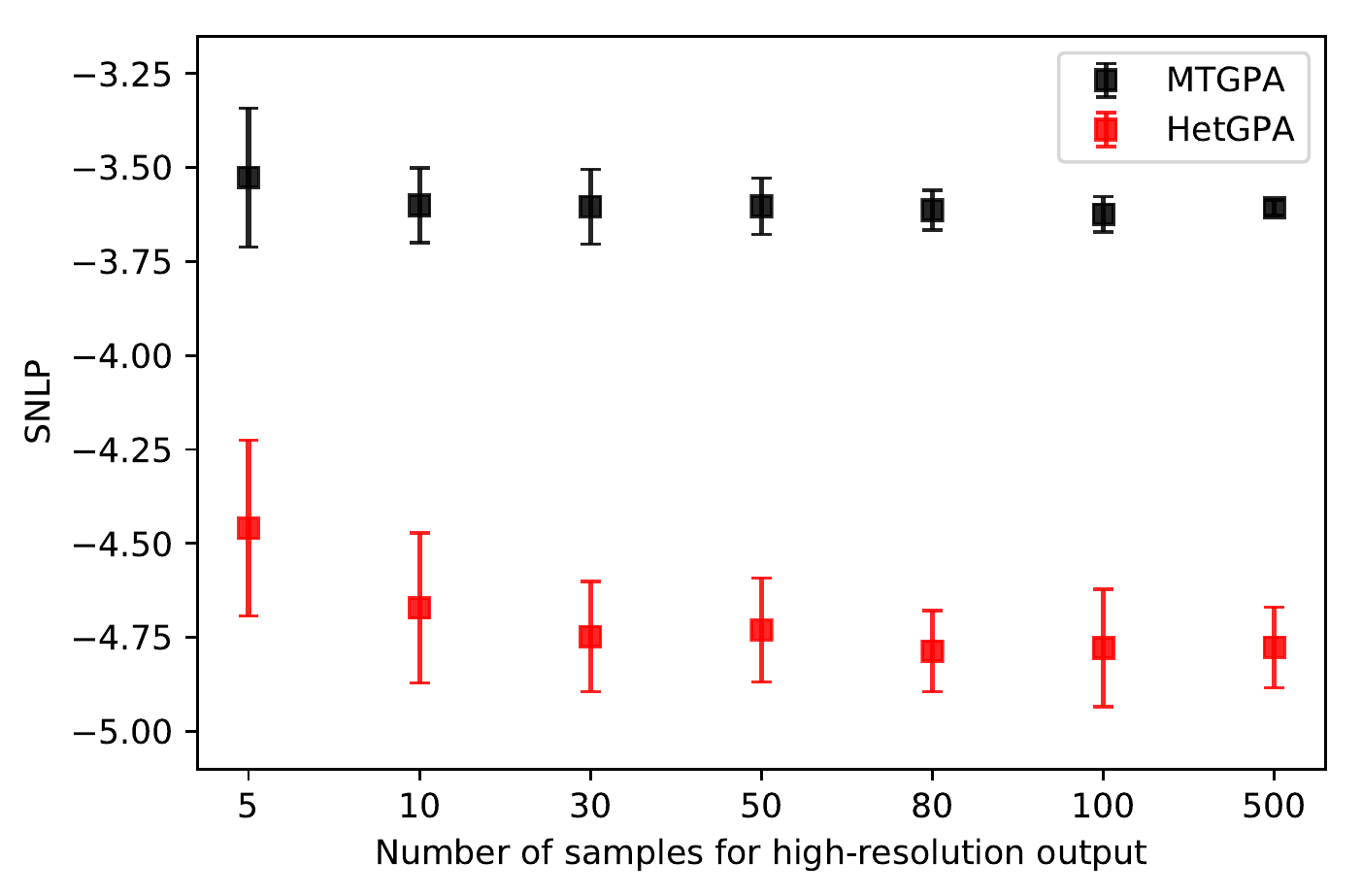}
	\caption{SNLP plots for the fertility dataset for $5\times 5$ (left panel) and $2\times 2$ (right panel) aggregated data for four outputs (two fertility rates). All outputs are considered as Gaussian (MTGPA) and all outputs are considered as heteroscedastic Gaussian (HetGPA).}
	\label{fig:snlp:hetgp}
\end{figure}

In Figure \ref{fig:snlp:hetgp} SNLP is calculated for four outputs (two outputs with high-resolution and a few data points and two
outputs with low-resolution and many more data points). The high-resolution data correspond to the fertility rates of the first
and second birth orders. 

The first task consists of a different number of data observations randomly taken from the training points of the fertility rate of the first birth. 
The second task consists of all the training data at the first task aggregated at two different resolutions, $5 \times 5$ and $2 \times 2$.

The third task consists of a different number of data observations randomly taken from the training points of the fertility rate of the second birth.  
The fourth task consists of all the training data at the third task aggregated at two different resolutions, $5 \times 5$ and $2 \times 2$.

We use two different versions of our model and compare their SNLPs. In one version, all the outputs are
considered as Gaussians (MTGPA) and in the second version, all the outputs are considered as heteroscedastic Gaussians (HetGPA). 
In the Gaussian case, only the mean parameter is modelled as a latent function, while the variance is a hyperparameter. However, in the heteroscedastic case, both mean and variance are assumed to follow latent functions. The model with HetGPA outperforms the model with MTGPA since it allows more flexibility toward the latent function that models the variance of the Gaussian.

\vspace{25mm}
\subsection{Bar plot for the synthetic dataset}
\begin{figure}[H]
	\centering
	\includegraphics[width=0.95\textwidth]{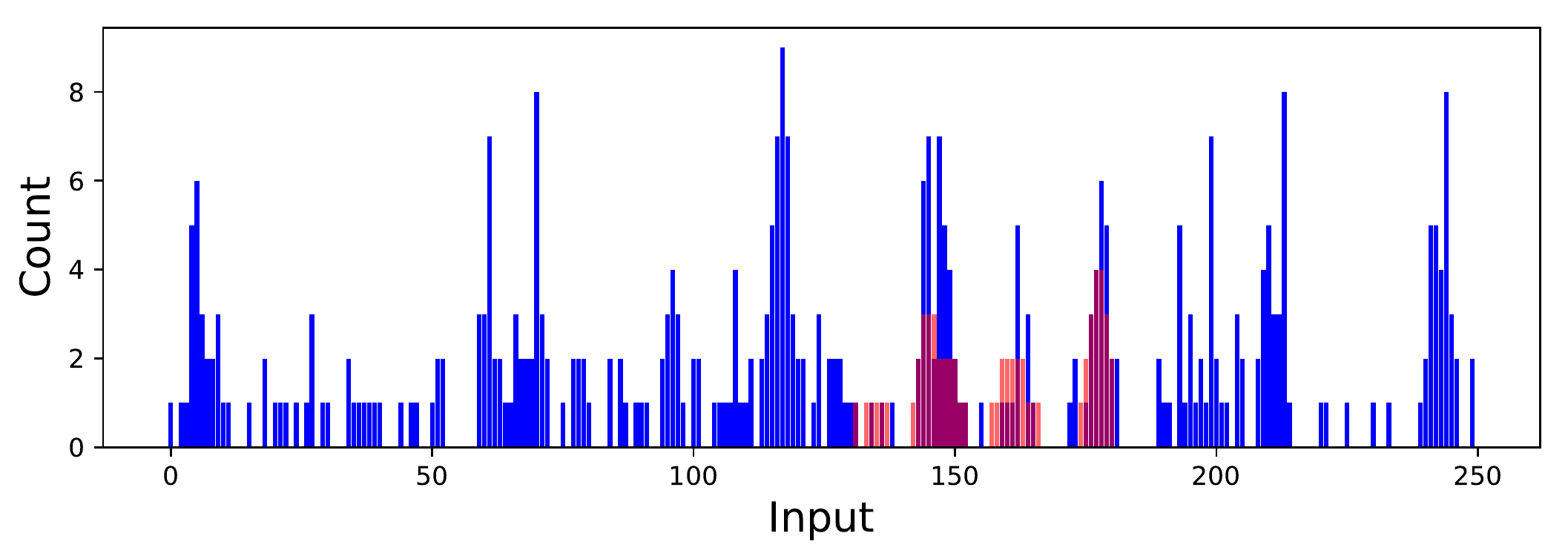}
	\caption{Counts for the Poisson likelihood and predictions for the multi-task model. Predictions are illustrated only for the first task (the one with support of $\upsilon_1=1$). The blue bars are the original one-unit support data, and the red bars are the predicted test results in the gap $[130, 180]$. We did not include the two-unit support data (the second task) for clarity in the visualisation.}
	\label{fig:errorbars:testonly}
\end{figure}

Figure \ref{fig:errorbars:testonly} shows the result for the synthetic count data with Poisson likelihood and prediction using the multi-task model. This plot is the same as Figure \ref{fig:toy}.b, however, the green bars are removed for better visualisation purposes.

\end{document}

%% file: synthetic.tex
In this section we evaluated our model with a synthetic dataset.  For all of the experiments we use $Q=1$ with an EQ covariance for the latent function $u_1(z)$.
We set up a toy problem with $D=2$ tasks, where both likelihood functions are Poisson. We sample from the latent vector-valued GP and use those samples
to modulate the Poisson likelihoods using $\exp(f_1(\cdot))$ and $\exp(f_2(\cdot))$ as the respective rates. The first task is generated using intervals of $\upsilon_1=1$ units, whereas the second task is generated using intervals of $\upsilon_2 = 2$ units. All the inputs are uniformly distributed in the range $[0, 250]$.
We generated $250$ observations for task 1 and $125$ for task 2. For training the multi-task model, we select $N_1 = 200$ from the $250$ observations for task 1 and use all $N_2=125$ for the second task. The other $50$ data points for task 1 correspond to a gap in the interval $[130, 180]$ that we use as the test set. In this experiment, we evaluated our model's capability in predicting one task, sampled more frequently, using the training information from a second task with a larger support. 

\begin{figure}
\centering
\begin{minipage}[b]{0.49\textwidth}
\includegraphics[width=\textwidth]{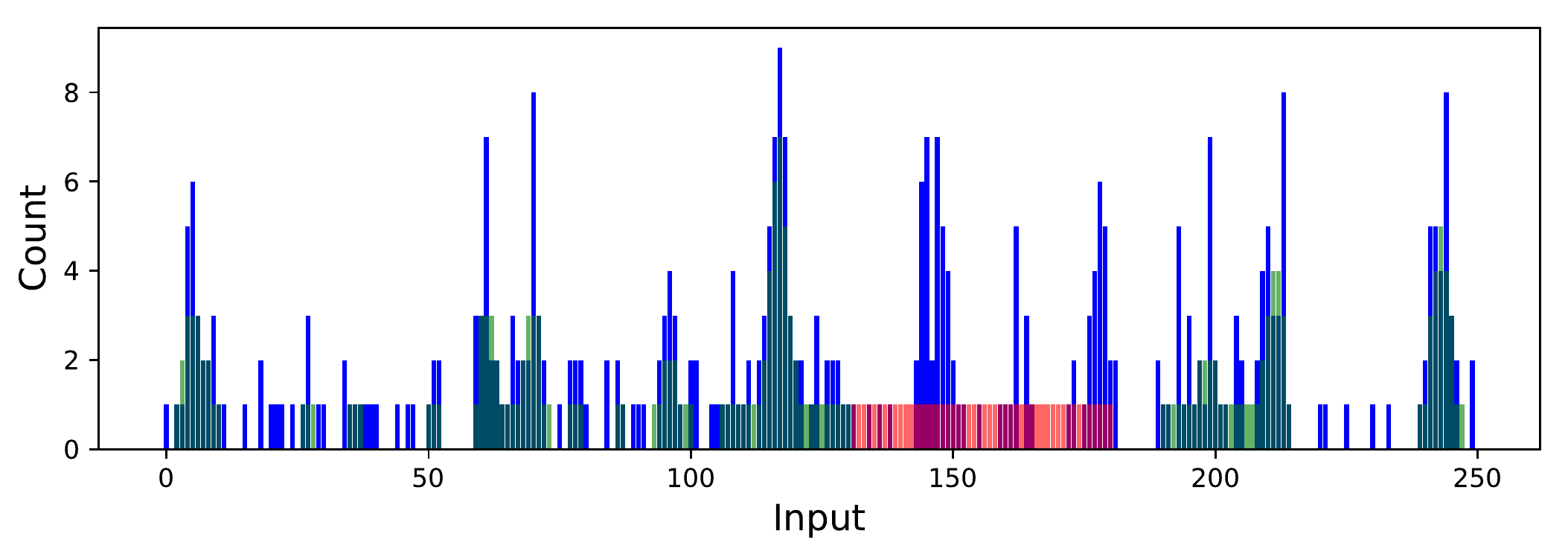}
\caption*{{\small{(a) Single-task learning}}}
\end{minipage}
\begin{minipage}[b]{0.49\textwidth}
\includegraphics[width=\textwidth]{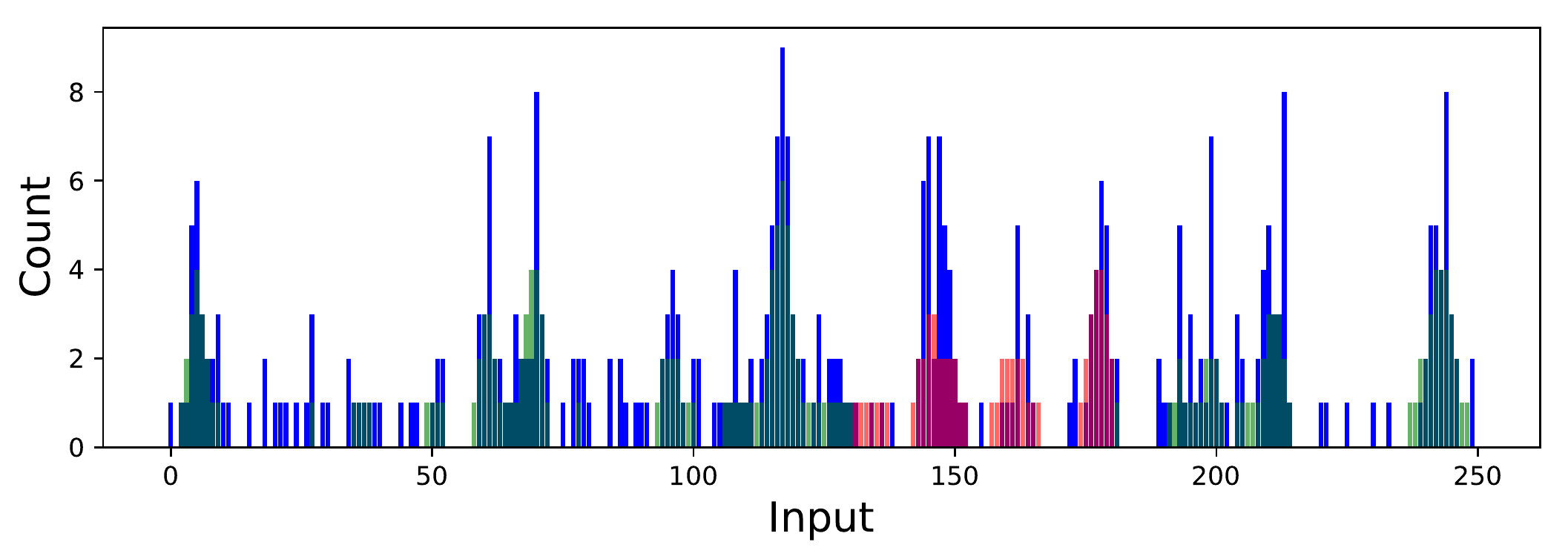}
\caption*{{\small{(b) Multi-task learning}}}
\end{minipage}
\caption{Counts for the Poisson likelihoods and predictions using the single-task vs multi-task models. Predictions are shown only for the first task (the one with support of $\upsilon_1=1$). The blue bars are the original one-unit support data, the green bars are the predicted training count data and the red bars are the predicted test results in the gap $[130, 180]$. We did not include the two-unit support
data (the second task) for clarity in the visualisation. The multi-task figure on the right (b) is illustrated again in Appendix \ref{appendix:sec:fertility:hetgp:toy}, Figure \ref{fig:errorbars:testonly} for better visualisation.}
\label{fig:toy}
\end{figure}

In Figure \ref{fig:toy} we show that the data in the second task, with a larger support, helps predicting the test data in the gap present in the first task, with a smaller support (right panel). However, this is not the case in the single task learning scenario where the predictions are basically constant and equal to 1 (left panel). Both models predict the training data equally well. SMSE (Standardized Mean Squared Error) and SNLP (standardized negative log probability density) are calculated for five independent runs.
For the multi-task scenario they are $0.464 \pm 0.136$ and $-0.822 \pm 0.109$ and for the single task case they are $0.9699\pm 0.016$ and $-0.095\pm 0.036$, respectively.

%% file: fertility.tex
In this experiment, a subset of the Canadian fertility dataset is used from the Human Fertility Database (HFD)
\footnote{\url{https://www.humanfertility.org}}.  
The dataset consists of live births' statistics by year, age of mother and birth order. 
The ages of the mother are between $[15, 54]$ and the years are between $[1944, 2009]$.
It contains $2640$ data points of fertility rate per birth order (the output variable) and there are four birth orders.  
We used the $2640$ data points of the 1st birth only. The dataset was randomly split into $1640$ training points and $1000$ test points. We consider two tasks: the first task consists of a different number of data observations randomly taken from the $1640$ training points. The second task consists of all the training data aggregated at two different resolutions, $5\times 5$ and $2\times 2$ (we wanted to test the predictive performance when the relation of high-resolution data to low-resolution data was $1^2$ to $5^2$ and another for $1^2$ to $2^2$).
The aggregated data for the $5\times 5$ case (a squared support of $5$ years for the input \texttt{age}
times $5$ years for the input \texttt{years} of the study) is reduced to $104$ data points and the aggregated data for $2\times 2$ case is reduced to $660$ points. 

\begin{figure}[ht!]
	\centering
	\includegraphics[width=0.45\textwidth]{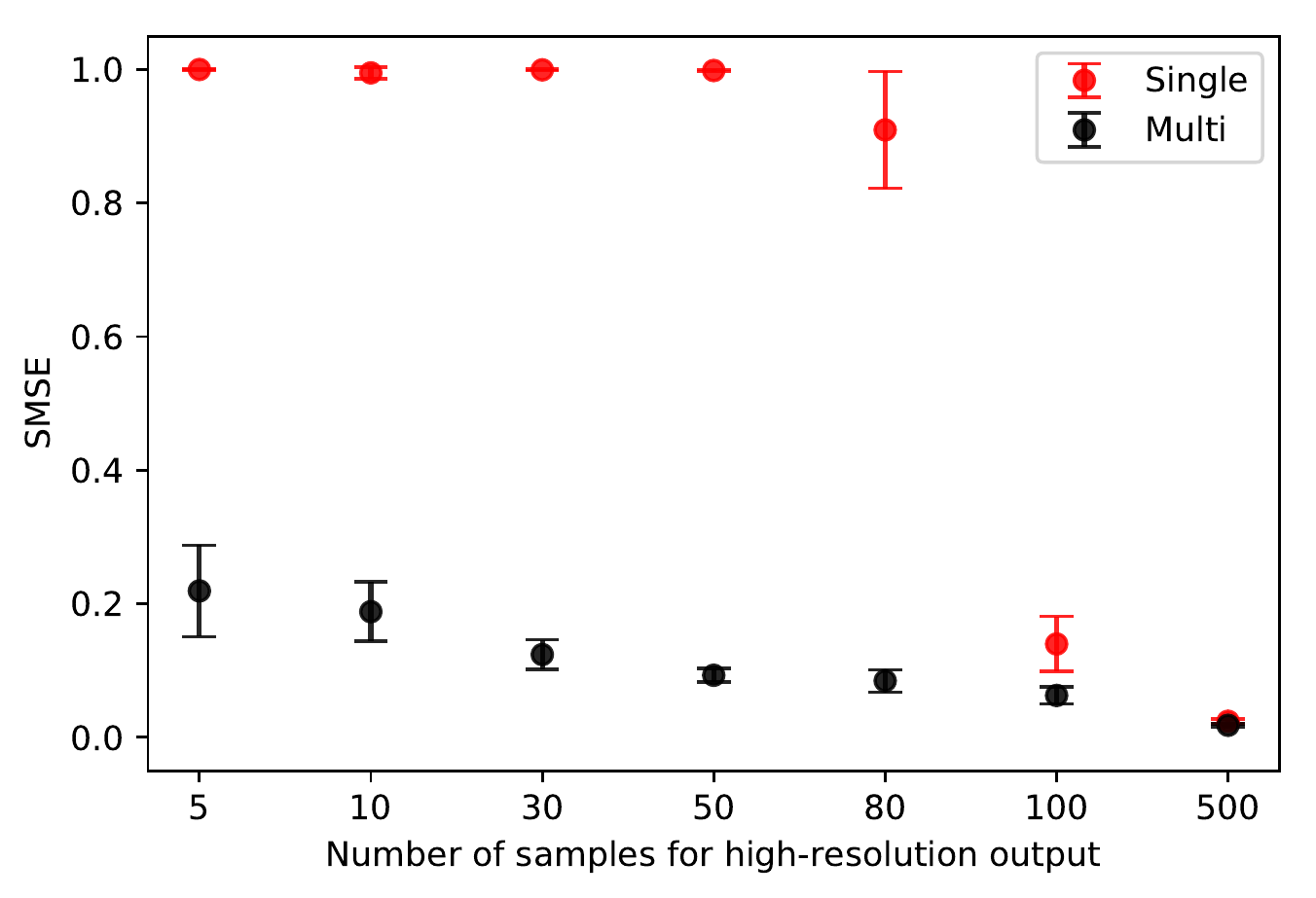}
	\includegraphics[width=0.45\textwidth]{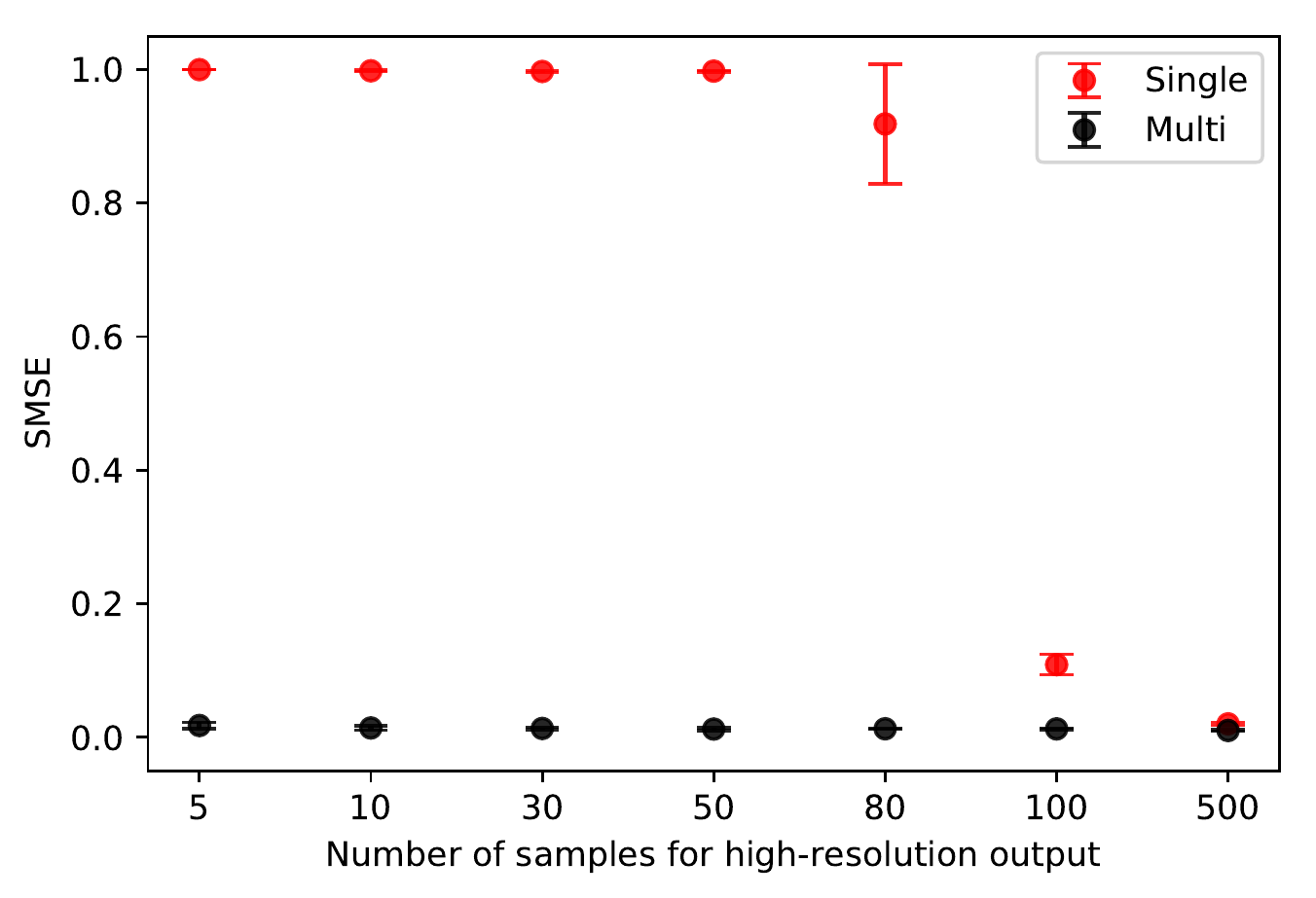}		
	\caption{SMSE plots for the fertility dataset for $5\times 5$ (left panel) and $2\times 2$ (right panel) aggregated data. 
		The Figure shows the performance in terms of the number of training instances used
		for the data sampled at a higher resolution. The test set always contains $1000$ instances. We plot the mean and standard deviation for five repetitions of the experiment with different sets of training
		and test data. Appendix
		\ref{appendix:sec:fertility:snlp} shows the same plots for SNLP.  Appendix \ref{appendix:sec:fertility:baselines} illustrates other experimental baselines that are compared to eachother for the same metrics.  Further experiments considering more tasks are also included in Appendix \ref{appendix:sec:fertility:hetgp:toy}.}\label{fig:fertility:smse}
\end{figure}

In the experiments, we train this multi-task model by slowly increasing the original resolution training data, while
maintaining a fixed amount of training points mentioned before for the aggregated second task. The output variable (fertility rate for the first birth) was assumed to be Gaussian, so both tasks follow a Gaussian likelihood. We use $Q=1$ with an EQ kernel $k_1(\boldz, \boldz')$ with $\boldz\in\mathbb{R}^2$ where the two input variables are age of mother and birth year. We used $100$ fixed inducing variables and mini-batches of size $50$ samples. 
The prediction task consists of predicting the $1000$ original resolution test data with the help of the second task which consists of the aggregated data ($5\times 5$ or $2\times 2$ for two separate experiments). 

Figure \ref{fig:fertility:smse} shows SMSE for five random selections of data points in the training and test sets. 
We notice that the multi-task learning model outperforms the single-task GP when there are few observations in the task with the original resolution data. This pattern holds below $500$ observations. 
At that point, both models perform equally well since the single-task GP now has enough training data. With respect to the two different resolutions, the performance of the multi-task model is better when the second task has a $2\times 2$ resolution rather than $5\times 5$ resolution, as one might also expect. 

%% file: airpollution.tex

   
Particulate air pollution can be measured accurately with high temporal precision by using a $\beta$ attenuation (BAM) sensor or similar. Unfortunately these are often prohibitively expensive. We propose instead that one can combine the measurements from a low-cost optical particle counter (OPC) which gives good temporal resolution but is often badly biased, with the results of a Cascade Impactors (CIs), which are a cheaper method for assessing the mass of particulate species but integrate over many hours (e.g. 6 or 24 hours). 

\begin{figure}[ht!]
	\centering
	\includegraphics[width=0.9\textwidth]{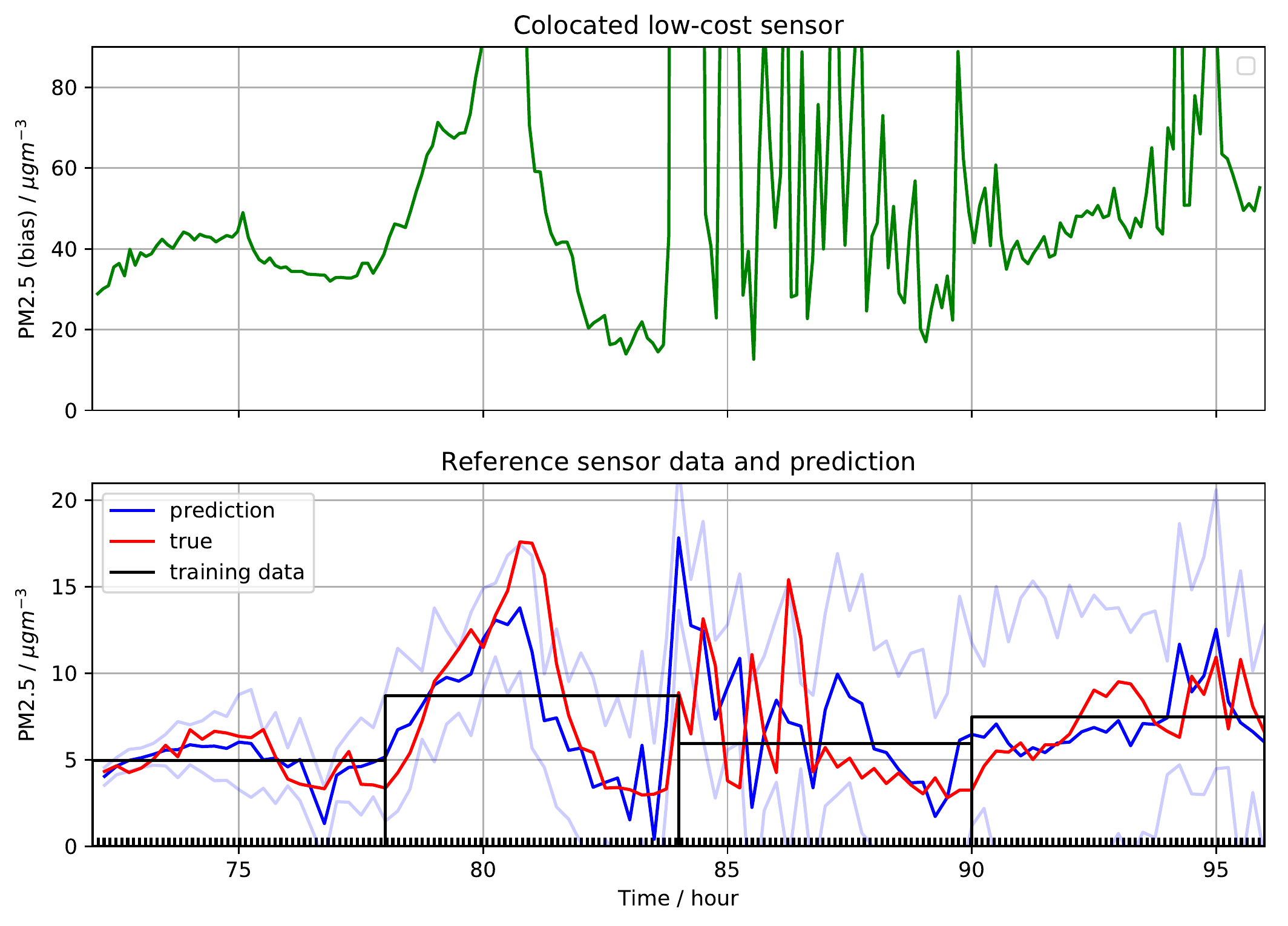}
	\caption{Upper plot: a (biased) OPC low-accuracy high-frequency measurement of PM2.5 air pollution. Lower plot: the high-precision low-frequency training data (black rectangles) the test data from the same instrument (red) and the posterior prediction for this output variable, predicting over the same 15-minute periods as the test data (blue, with pale blue indicating 95\% confidence intervals). The ticks in the bottom of the lower plot indicate the position of the inducing inputs. Also, we have deliberately cut the higher peaks of the samples in the upper plot that can go as high as 500 $\mu g/ m^3$, just to be able to visualise better the samples in other parts of the plot. \label{pollutionfig}}
\end{figure}

One can formulate the problem as observations of integrals of a latent function. The CI integrating over 6 hour periods while the OPC sensor integrating over short 5 minute periods. We used data from two fine particulate matter (PM) sensors. The sensors are less than $2.5$ micrometer diameter (PM2.5) and are colocated in Kampala, Uganda at 0.3073${}^\circ$N 32.6205${}^\circ$E. The data is taken between 2019-03-13 and 2019-03-22. We used the average of six-hour periods from a calibrated mcerts-verified Osiris (Turnkey) particulate air pollution monitoring system as the low-resolution data, and then compared the prediction results to the original measurements (available at a 15 minute resolution). We used a PMS 5003 (Plantower) low-cost OPC to provide the high-resolution data. Typically we found these values would often be biased. We simply normalised (scaled) the data to emphasise that the \emph{absolute} values of these variables are not of interest in this model.

Our multi-task model consists of a single latent function, $Q=1$, with covariance $k_1(z,z')$ that follows an EQ form. We assume both outputs follow Gaussian likelihoods. In our model, one task represents the high accuracy low-resolution samples and the second task represents the low-accuracy high-resolution samples. The posterior GP both aims to fulfil the 6-hour long integrals of the high-accuracy data (from the Osiris instrument) while remaining correlated with the high-frequency bias data from the OPC. We used 2000 iterations of the variational EM algorithm, with 200 evenly spaced inducing points and a fixed lengthscale of 0.75 hours. We only optimise the parameters of the coregionalisation matrix $\boldB_1\in\mathbb{R}^{2\times 2}$ and the variance of the noise of each Gaussian likelihood.

Figure \ref{pollutionfig} illustrates the results for a 24 hour period. The training data consists of the high-resolution low-accuracy sensor and a low-frequency high accuracy sensor. The aim is to reconstruct the underlying level of pollution both sensors are measuring. To test whether the additional high-frequency data improves the accuracy we ran the coregionalisation both with and without this additional training data.

We found that the SMSE for the predictions over the 9 days tested were substantially smaller with multi-task learning compared to using only the low-resolution
samples,  $0.439\pm0.114$ and $0.657\pm0.100$ respectively (the difference is statistical significant using a paired $t$-test with a $p$ value of $0.0008$). In summary, the model was able to incorporate this additional data and use it to improve the estimates while still ensuring the long integrals were largely satisfied.
